\documentclass[letterpaper, 10 pt, conference]{ieeeconf}  % Comment this line out if you need a4paper

\IEEEoverridecommandlockouts                              % This command is only needed if 
\usepackage[absolute,overlay]{textpos}
\usepackage{graphicx}   % 그림 삽입
\usepackage{tabularx}
\usepackage{wrapfig}    % 본문 흐름과 함께 감싸기
\usepackage{mwe}        % example-image 계열 더미 그림
\usepackage{lipsum}     % 예시용 더미 텍스트
\usepackage{amsmath}
\usepackage{amssymb}
\usepackage{tikz}
\usepackage{tikzpagenodes}
\usepackage{booktabs}
\usepackage{multirow}
\usepackage[table]{xcolor}
\usepackage{etoolbox}
\usepackage{authblk}
\usepackage{hyperref}
\usepackage[dvipsnames]{xcolor}

\usepackage{caption}
\usepackage{float}
\usepackage{authblk}
\usepackage{cite} % tm - 이거 추가함
\usepackage{booktabs,multirow,siunitx}
\usepackage{fancyhdr}
\fancypagestyle{firstpage}{
  \fancyhf{}
  \fancyfoot[C]{\footnotesize Preprint. Accepted at IEEE International Conference on Robotics and Automation (ICRA), 2026.}

}

\newcommand{\ms}[2]{\num{#1 \pm #2}}
\makeatletter
\newcommand{\setupitem}[2]{\noindent\textbf{#1}\ #2\par\vspace{2pt}}
\patchcmd{\@makecaption}{\MakeUppercase}{\relax}{}{}
\patchcmd{\@makecaption}{\scshape}{\normalfont}{}{}
\makeatother
\title{\LARGE \bf
LEGO: Latent‑space Exploration for Geometry‑aware \\ Optimization of Humanoid Kinematic Design
}

\author{Jihwan Yoon$^{1}$, Taemoon Jeong$^{1}$, Jeongeun Park$^{1}$, Chanwoo Kim$^{1}$,\\
Jaewoon Kwon$^{2}$, Yonghyeon Lee$^{3}$, Kyungjae Lee$^{4}$, and Sungjoon Choi$^{1}\textsuperscript{*}$% <- 저자 명단 끝
\thanks{This work was supported by Institute of Information \& communications Technology Planning \& Evaluation (IITP) grant funded by the Korea government(MSIT): (No. RS-2024-00457882, AI Research Hub Project, 50\%), (No. RS-2022-II220871, Development of AI Autonomy and Knowledge Enhancement for AI Agent Collaboration, 25\%), Ministry of Trade, Industry, and Energy (MOTIE), Korea, under the “Global Industrial Technology Cooperation Center program” supervised by the Korea Institute for Advancement of Technology (KIAT).  (Grant No. P0028435, 25\%).}
\thanks{$^{1}$Jihwan Yoon, Taemoon Jeong, Jeongeun Park, Chanwoo Kim, and Sungjoon Choi are with the Department of Artificial Intelligence, Korea University, Seoul 02841, Republic of Korea (e-mail: \{yoonmungchi, taemoon-jeong, baro0906, chanwoo-kim, sungjoon-choi\}@korea.ac.kr).}%
\thanks{$^{2}$Jaewoon Kwon is with Contoro Robotics, Austin, TX 78758, USA (e-mail: jaewoon@contoro.com).}%
\thanks{$^{3}$Yonghyeon Lee is with the Department of Mechanical Engineering, Massachusetts Institute of Technology, Cambridge, MA 02139, USA (e-mail: yhl@mit.edu).}%
\thanks{$^{4}$Kyungjae Lee is with the Department of Statistics, Korea University, Seoul 02841, Republic of Korea (e-mail: kyungjae\_lee@korea.ac.kr).}%
\thanks{$^{*}$Corresponding author.}% <- 교신저자 표시 추가 (선택사항)

}

\begin{document}
\setlength{\abovedisplayskip}{3pt}
\setlength{\belowdisplayskip}{3pt}
% \linespread{0.987}
\bstctlcite{IEEEexample:BSTcontrol}

\maketitle
% \maketitle 뒤에:
\thispagestyle{firstpage}
\pagestyle{empty}

%%%%%%%%%%%%%%%%%%%%%%%%%%%%%%%%%%%%%%%%%%%%%%%%%%%%%%%%%%%%%%%%%%%%%%%%%%%%%%%%
\begin{abstract}
Designing robot morphologies and kinematics has traditionally relied on human intuition, with little systematic foundation. Motion–design co-optimization offers a promising path toward automation, but two major challenges remain: (i) the vast, unstructured design space and (ii) the difficulty of constructing task-specific loss functions. We propose a new paradigm that minimizes human involvement by (i) learning the design search space from existing mechanical designs, rather than hand-crafting it, and (ii) defining the loss directly from human motion data via motion retargeting and Procrustes analysis. Using screw-theory-based joint axis representation and isometric manifold learning, we construct a compact, geometry-preserving latent space of humanoid upper body designs in which optimization is tractable. We then solve design optimization in this latent space using gradient-free optimization. Our approach establishes a principled framework for data-driven robot design and demonstrates that leveraging existing designs and human motion can effectively guide the automated discovery of novel robot design. Project page: \url{https://jihwan-yoon-page.github.io/legopt/}
\end{abstract}

%%%%%%%%%%%%%%%%%%%%%%%%%%%%%%%%%%%%%%%%%%%%%%%%%%%%%%%%%%%%%%%%%%%%%%%%%%%%%%%%
\section{Introduction}

In robot mechanical design, determining the morphology as well as the lengths of links and axes of joints is typically the very first step. In practice, however, this process often depends heavily on human intuition and lacks a principled, systematic foundation. To move toward design automation, a promising approach is motion–design co-optimization~\cite{kwon2022physically}, which, given a set of target motions, seeks designs that are inherently suited to executing those motions -- a direction to which this paper contributes.

The major difficulty lies in the vast, combinatorial, and unstructured nature of the design space. To make optimization tractable, designers must first restrict the search space -- for example, by exploiting task-specific heuristics, symmetry assumptions, or low-dimensional parametric families -- before meaningful optimization can be performed~\cite{kwon2022physically}. Moreover, each new task often introduces unique constraints and parameters, necessitating repeated and costly redefinition of the design space.

Another challenge lies in designing the loss (or reward) function for a given task. For example, given target robot behaviors (e.g., a specific style of dancing), defining an appropriate loss or reward is nontrivial. Even when possible, it typically requires many iterations, significant human involvement, and substantial time and computational resources.

In this paper, we propose a new paradigm that minimizes human involvement. The main idea is twofold. First, instead of being hand-crafted, the search space is learned from existing mechanical designs used as training data. Although the available designs do not constitute a massive dataset on the scale of vision or language data, they are nevertheless diverse enough to capture meaningful structural patterns and priors for guiding design optimization -- a point we demonstrate in this paper.

Second, the loss function is defined directly from human motion data, requiring data but not heuristic or costly manual design. In this regard, we address the key challenge of reconciling the discrepancy between the human skeletal kinematic structure and the robot kinematic structure being optimized by leveraging motion retargeting techniques and Procrustes analysis.

Regarding the search space, the first question that naturally arises is how to numerically represent existing designs. Adopting screw theory~\cite{lynch2017modern, park1994computational}, we model each joint as a 6D vector $(\boldsymbol{\omega}, \mathbf{q})$, with $\boldsymbol{\omega}$ a unit axis and $\mathbf{q}$ the joint position in a common base frame. We focus on designs with broadly similar morphologies -- such as humanoids with conventional kinematic trees -- so that a design with $N$ joints corresponds to a point in $\mathbb{R}^{6N}$, with padding applied for those with fewer joints.\looseness = -1

The next question is how to identify a lower-dimensional subspace of existing designs that enables efficient optimization. Although each design resides in a high-dimensional space of size $6N$ (e.g., humanoids with over 20 joints correspond to more than 100 dimensions), the set of realizable designs is expected to lie on a structured, lower-dimensional manifold. To capture this, we use an encoder–decoder framework for manifold learning~\cite{lee2023geometric, lee2021neighborhood, lee2023explicit, lee2022regularized}, mapping designs into a compact, lower-dimensional yet expressive latent space.

To further improve structure in the latent space, we employ an isometric regularization approach, which encourages smoothness and geometry preservation~\cite{lee2022regularized,lim2024graph, lee2024mmp++,heo2025isometric}. 
Optimization is then carried out directly in this latent space, with the loss function defined from human motion data -- via Procrustes analysis -- and solved using Voronoi Optimistic Optimization (VOO)~\cite{kim2020voot}.

The contributions of this paper are summarized as follows:
\begin{itemize}
\item A screw-theory-based representation that enables intuitive and scalable encoding of kinematic structures.  
\item A learning-based framework for search space design that leverages existing mechanical designs.  
\item An integrated pipeline that combines an isometric autoencoder, a human-motion–retargeting-based loss function, and VOO.  
\end{itemize}

%The remainder of the paper is structured as follows. Section II reviews related literature. Section III describes screw-theory-based data representation and dataset creation. Section IV discusses the isometric autoencoder and latent space learning. Section V outlines latent space optimization using VOO. Finally, Section VI provides experimental validation and comparisons against alternative representation methods.

\begin{figure*}[!t]                % !t: 가능한 Top 선호
  \centering
  \includegraphics[width=\linewidth]{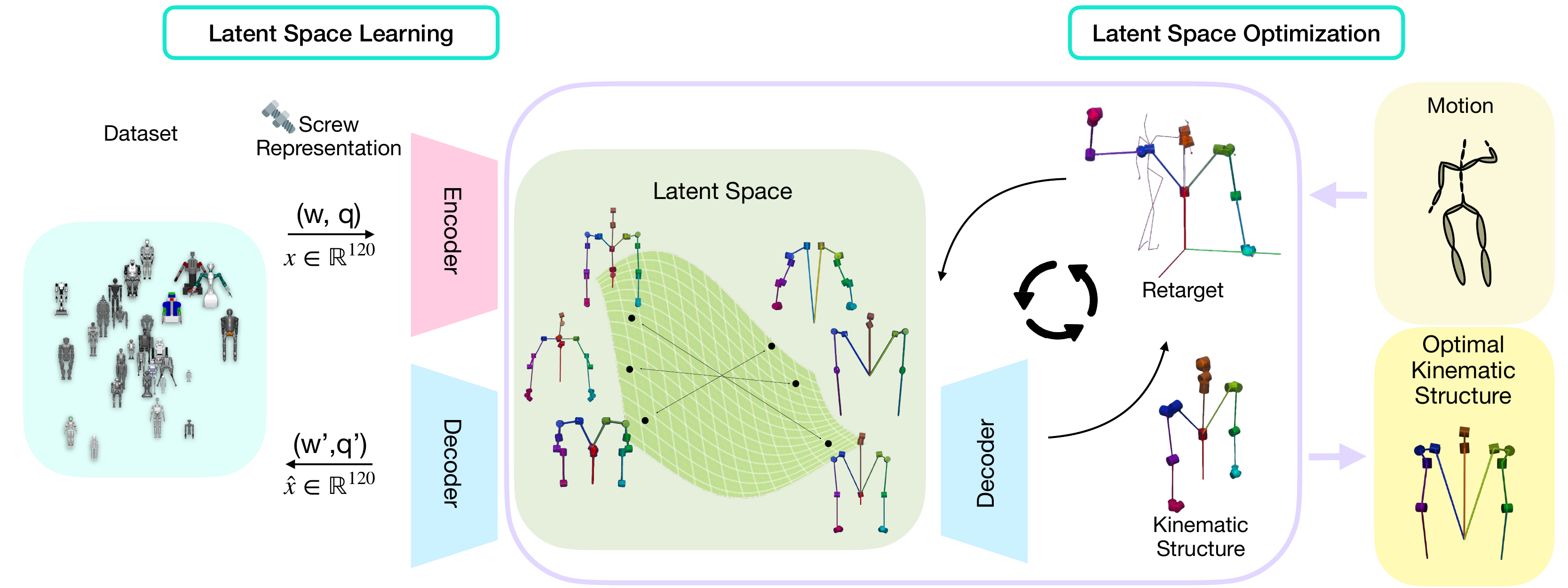} % 컬럼폭 사용
  % \caption{Total pipeline (blueprint)}
  \caption{Total pipeline for humanoid kinematic structure optimization. First, a dataset of robots is converted to a unified \textbf{Screw Representation} ($(\omega, q)$). An autoencoder with an \textbf{isometric regularization loss} learns a compact 2D latent space from these representations. Next, \textbf{Voronoi Optimistic Optimization (VOO)} iteratively samples and decodes candidates from the latent space. Each candidate is evaluated by its motion retargeting performance using \textbf{Procrustes analysis}. The process converges to an optimal kinematic structure that best performs the given task.}
  \label{fig:fullwidth}
  % \vspace{-3mm}
\end{figure*}

\section{Related Work: Learning-based Design Optimization}
%$%%%%%%%%%%%%%%
% Related works
%$%%%%%%%%%%%%%%
% \subsection{Screw-theoretic Representations.} Screw theory provides a compact and expressive way to describe rigid body motion through six-dimensional twist and wrench vectors, capturing both angular and linear components in a unified form~\cite{ball1900}. While it has been widely adopted in robot kinematics and control, its application in robot design representation remains limited. Recent works such as ScrewMimic~\cite{bahety2024screwmimic} have shown its utility in modeling coordinated bimanual manipulation by projecting human hand motions into a screw-action space. These approaches demonstrate that screw-theoretic formulations can provide geometric structure and consistency, especially when reasoning about movement across diverse morphologies. In contrast to prior work focusing on control or manipulation, our approach utilizes screw theory to construct a representation space specifically for humanoid upper body morphology, serving as the foundation for learning and optimization. 

In this section, we review learning-based design methods, focusing on two main directions: latent-model–based optimization~\cite{hu2023glso,bentley2022coil,song2024morphvae}, which is closely related to our framework, and reinforcement learning approaches~\cite{he2024morph,schaff2022nlimb}.

To address the curse of dimensionality in robot design, some recent approaches propose learning low-dimensional latent spaces that retain design validity while enabling efficient optimization~\cite{hu2023glso,bentley2022coil,song2024morphvae}. 
Grammar-guided Latent Space Optimization (GLSO)~\cite{hu2023glso} embeds modular robot structures into a continuous latent space using a graph VAE, allowing Bayesian optimization to be performed with high sample efficiency. 
COIL~\cite{bentley2022coil} constrains the latent space to represent only valid designs and uses evolutionary strategies to search it. MorphVAE~\cite{song2024morphvae} applies VAE-based learning to soft robot morphology using continuous evolutionary sampling between tasks and generations. 

Other methods, such as MORPH~\cite{he2024morph} and N-LIMB~\cite{schaff2022nlimb}, integrate reinforcement learning with differentiable or universal control policies to co-optimize morphology and behavior. However, these approaches are computationally inefficient, lack search-space reduction, and often explore spaces so large that they lead to designs that are not practically manufacturable.

Our approach differs in three major ways:
(i) prior works rely on graph or voxel representations, whereas we adopt a screw-theory-based representation;
(ii) they depend on synthetic datasets generated under human-defined constraints, whereas we leverage real, existing robot designs; and
(iii) their loss functions are manually designed, while ours is directly defined from human motion data.
Taken together, these distinctions advance our framework toward more automated and less human-involved robot design. Additionally, while all of these methods remain limited to simulation studies, we present a real robot design, demonstrating the real-world applicability of our approach.

% These are latent + optimization
% Data voxel -> representation에 대한 이야기
% synthetic data generation (constraints, rejection)
% representation 다르고
% synthetic data (some hueristic involved) vs real data
% design loss function (demonstration data) -> loss 설계

% RL? Learnig-based. -> Reward design, computation efficiency.
% Other methods such as MORPH~\cite{he2024morph} and N-LIMB~\cite{schaff2022nlimb} integrate reinforcement learning with differentiable or universal control policies to co-optimize morphology and behavior. 

% All of them are simulation studies no real design.
% These approaches largely rely on VAE priors or differentiable simulations and often target modular or voxel-based morphologies. In contrast, our work focuses on humanoid upper-body structures and leverages an isometrically regularized autoencoder to preserve local geometry in the latent space, enabling optimization using Voronoi-based sampling methods.

\section{Preliminaries}
Our framework consists of three main components: (i) representation of the kinematic structure, (ii) data-driven dimensionality reduction of the search space, and (iii) optimization via retargeting in the reduced space. To support these components, this section introduces screw theory, manifold learning, and motion retargeting with Procrustes analysis, which together serve as the fundamental building blocks of our design framework.
%$%%%%%%%%%%%%%%
% Prelim.
%$%%%%%%%%%%%%%%
% 1. Screw theory
% 2. Procrustes analysis -> loss
% 3. Manifold (latent) learning + Isometric regularization
\subsection{Screw Theory}
According to the Chasles–Mozzi theorem, any rigid-body displacement can be expressed as a screw motion, i.e., a rotation combined with a translation along a fixed line in space called the screw axis~\cite{murray2017mathematical, lynch2017modern, park2018geometric}.
The screw axis is defined as the line passing through a point $\mathbf{q}$ and oriented by a unit vector $\boldsymbol{\omega}$. The displacement consists of a rotation about $\boldsymbol{\omega}$ by an angle $\theta$, together with a translation parallel to $\boldsymbol{\omega}$ of magnitude $\theta h$, where $h$ denotes the screw pitch.
Then, the corresponding rigid-body motion is characterized by this screw motion:
\begin{equation}
    \exp \big(\begin{bmatrix}
        [\boldsymbol{\omega}] & \mathbf{v} \\ 
        0 & 0
    \end{bmatrix}\big) \in {\rm SE}(3) \subset \mathbb{R}^{4\times 4},
\end{equation}
where 
\begin{equation}
    [\boldsymbol{\omega}] = \begin{bmatrix}
        0 & -\omega_3 & \omega_2 \\
        \omega_3 & 0 & -\omega_1 \\
        -\omega_2 & \omega_1 & 0
    \end{bmatrix}
\end{equation}
and $\mathbf{v}=-\boldsymbol{\omega} \times \mathbf{q} + h\boldsymbol{\omega}$. 
Thus, $(\boldsymbol{\omega}, h, \mathbf{q})$ provides a complete parametrization of a rigid-body motion via its screw axis: $\boldsymbol{\omega}$ specifies the axis direction, $h$ the screw pitch, and $q$ a point on the axis.
In particular, the case $h=0$ corresponds to pure rotation, which we use as a compact geometric representation of each joint axis throughout the paper to effectively represent robot models.

\subsection{Manifold Learning and Isometric Regularization}
We view an autoencoder, consisting of an encoder $f$ that maps a data point $\mathbf{x}$ to a lower-dimensional latent variable $\mathbf{z}$ and a decoder $g:\mathbf{z}\mapsto \mathbf{x}$, as learning a chart of the data manifold~\cite{lee2023geometric,lee2021neighborhood}, i.e., the decoder $g$ parametrizes the data manifold.
The pullback metric on the latent space is given by $G(\mathbf{z})=J_g(\mathbf{z})^\top J_g(\mathbf{z})$, where $J_g$ denotes the Jacobian of $g$.
The deviation of $G(\mathbf{z})$ from the identity matrix $\mathbf{I}$ quantifies local geometric distortion (to see why, consider $d\mathbf{x} = J_g(\mathbf{z})d\mathbf{z}$ and $d\mathbf{x}^\top d\mathbf{x} = d\mathbf{z}^\top J_g^\top J_g d\mathbf{z}$).

Isometry regularization encourages the latent space to preserve the geometry of the data manifold by reducing distortion~\cite{lee2022regularized,lim2024graph,heo2025isometric}, i.e., encouraging $J_g^\top J_g\approx \alpha \mathbf{I}$ (isometry up to scale), and has also been shown to yield a (intrinsically) flatter manifold~\cite{lee2023explicit}.
Specifically, the \emph{isometric} regularization term is derived as
\begin{equation}
    \mathcal{L}_{\rm iso} = \frac{\mathbb{E}_z[{\rm Tr}((J_g^\top J_g)^2)]}{\mathbb{E}_z[{{\rm Tr}(J_g^\top J_g)}]^2},
\end{equation}
where $\mathbb{E}_z$ is over latent space data distribution, which in practice can be efficiently computed by using Jacobian-vector and vector-Jacobian products~\cite{lee2022regularized}.
Isometric regularization stabilizes training with limited data and yields a well-conditioned, geometry-preserving latent design space, along with a smooth manifold well suited for our kinematic optimization.\looseness = -1

\subsection{Motion Retargeting and Procrustes Analysis}
We follow a motion retargeting strategy inspired by prior works on humanoid motion transfer~\cite{choikim,jeong2025core,jeong2025robust}. Because direct joint mapping is generally infeasible due to morphological differences, we define the target positions of the robot’s joints of interest (JOI) using scaled directional vectors from the human skeleton. The corresponding joint angles are then computed by solving a numerical inverse kinematics problem under joint limit constraints. To quantify the similarity between the source and retargeted motions, we employ Procrustes analysis~\cite{goodall1991procrustes}, an alignment method that removes differences in translation, rotation, and scale between trajectories. This yields the Procrustes Aligned Mean Per Joint Position Error (PA-MPJPE)~\cite{li2021hybrik,liu2021normalized,gartner2022differentiable,yang2023synbody,gartner2022trajectory}, which serves as a normalized measure of motion similarity and provides the basis for evaluating candidate kinematic structures in our optimization framework.

\section{Method}
%%%%%%%%%%%%%%%%%%%%%%%%%
% Methods
%%%%%%%%%%%%%%%%%%%%%%%%%
%
% In the following sections, ..... 
% we introduce problem defineition in LEGO.1.. 
% 
%
%%%%섹션에 대한 간략한 소개 적기(in this section, we'll talk about~). 문단이 5개이기 때문에 필요.
In this section, we present our pipeline for motion-specific humanoid upper body design optimization. 
We begin by formulating the problem as a structure–motion co-optimization task, where the goal is to identify a kinematic structure that minimizes the discrepancy between a target human motion and its retargeted execution on the robot.
We then introduce a screw-theoretic representation $(w,q)$ to encode upper body humanoid structures in global coordinates, followed by a data curation process that unifies 30 diverse robot models into a consistent 120-dimensional feature vector. 
Next, we learn a compact latent design space using an isometric-regularized autoencoder, which preserves geometric continuity while reducing dimensionality. 
Finally, we perform optimization in this latent space with VOO, combining motion retargeting and Procrustes analysis to identify feasible, symmetric upper-body designs tailored to specific motions.
\subsection{Problem Formulation}

We are given a source motion $M_{\text{src}}$, represented as a set of joint trajectories captured from a source kinematic model (e.g., a human skeleton). Our objective is to identify a new kinematic structure $\mathbf{x} \in \mathcal{X}$, together with its retargeted motion, such that the retargeted motion closely matches $M_{\text{src}}$. Here, $\mathcal{X}$ denotes an abstract design space of admissible upper-body kinematic structures, whose precise instantiation will be specified later.

To this end, we define a retargeting operator $M_{\rm tar} = \mathcal{R}(\mathbf{x};\,M_{\text{src}})$ which maps the source motion $M_{\mathrm{src}}$ onto a candidate kinematic structure $\mathbf{x}$, yielding the corresponding retargeted motion $M_{\mathrm{tar}}$ of $\mathbf{x}$.
The operator $\mathcal{R}$ is treated as a black-box procedure: it does not require differentiability and, in practice, is not differentiable due to the discrete nature of joint mappings and inverse kinematics. The optimization problem is thus formulated as

\begin{equation}
\mathbf{x}^* = \arg\min_{\mathbf{x} \in \mathcal{X}} \; \mathcal{L} \Big(M_{\text{src}}, \; \mathcal{R}(\mathbf{x};\,M_{\text{src}})\Big),
\end{equation}
where the loss function $\mathcal{L}$ measures the discrepancy between the source motion and the retargeted motion; in our experiments, we instantiate $\mathcal{L}$ using Procrustes analysis~\cite{Gower1975GeneralizedProcrustes}.
Given the optimal kinematic structure $\mathbf{x}^*$, its corresponding retargeted motion $M^*$ is 
$
M^* = \mathcal{R}(\mathbf{x}^*;\,M_{\text{src}}).$

This general formulation captures the essence of our goal: to search over possible kinematic structures in $\mathcal{X}$ and evaluate them by how well the retargeted motion aligns with the original source motion. In the following sections, we describe how this problem is addressed efficiently by introducing a learned latent design space and performing optimization therein.

% problem formulation 이후 screw theory-based representation에서 x를 representation으로 표현하기 이전에 이야기가 나와야 할 것 같다(formulate -> latent space 위에서 kinematic structure space를 정의 -> 그걸 위한 data representation). CMU Mocap 관련 이야기는 experiment로 넘기기.
% The latent space \( \mathcal{X} \) is obtained by training the autoencoder on existing upper-body humanoid structures encoded via screw-theoretic representations. Evaluation motions are taken from the CMU Mocap dataset~\cite{cmumocap2007}.

\subsection{Screw Theory-based Representation}
%%% 2. Screw-based humanoid representations
% joint number is not the same
% for learning latent model in later section, we need to match the data size
% challanging points
% revolute jonts usage ->
% maximum 30 joints to 8 joints
% padding: "w = (0, 0, 0)", q = (neighborhood joints aveage -> reasonable)  
% - joint가 없는 애들은 w가 어떤 같은 constant "c" 을 가져야한다. 
% - c랑 임의의 axis, unit vector, w랑 거리를 측정하면, 거리가 다 같아야한다.
% Good -> but, high-dimensional -> optimizaiton difficult, regulariztion neeed
%%%%
% "werid results" -> report ??
%%%

% Screw theory has been applied in various robotic contexts to unify rotation and translation through a compact, globally consistent formulation~\cite{bahety2024screwmimic}. In our context, designing for humanoid kinematics, \textbf{ positional fidelity} is paramount. We therefore opt for the representation:

% \begin{equation}
% S = (w, q), \quad w, q \in \mathbb{R}^3,
% \end{equation}%
% where $w$ is the global joint axis and $q$ the global joint position. 
% This preserves explicit positional information, enabling accurate retargeting and structural normalization across diverse morphologies. The traditional $(w, v)$ form, where $v$ is the moment vector, does not maintain spatial layout directly. 
% Nonetheless, if needed, the moment vector can be reconstructed using $v = -w \times q$.

% This $(w,q)$ formulation yields intuitive, scalable encoding of humanoid upper-body structures and, to our knowledge, marks the first such use for design-level humanoid optimization.

Screw theory has been widely used to unify rotation and translation within a compact, globally consistent formulation~\cite{lynch2017modern}. For our goal of comparing and retargeting motions in various upper body morphologies, global joint level positional fidelity is essential: the traditional form $(\boldsymbol{\omega},\mathbf{v})$, where $\mathbf{v}$ is the moment vector, does not directly encode absolute joint positions and depends on the choice of reference point, which can hinder spatial normalization and cross-morphology comparisons.

We therefore represent each joint by
$
    S=(\boldsymbol{\omega},\mathbf{q}), \quad \boldsymbol{\omega},\mathbf{q}\in\mathbb{R}^3,
$
where $\boldsymbol{\omega}$ denotes the direction (unit vector) of the global joint axis and $\mathbf{q}$ a point on that axis given in the world frame. This $(\boldsymbol{\omega},\mathbf{q})$ parameterization preserves explicit positional information, which is advantageous for motion retargeting and structural normalization. When needed (e.g., for revolute joints), the moment vector can be recovered as $\mathbf{v}=-\,\boldsymbol{\omega}\times \mathbf{q}$.

A practical challenge is that robots differ in the number of joints per anatomical group, whereas the learning stage requires fixed‑length inputs. To avoid injecting artificial directional bias during padding, we assign an orientationless placeholder to missing axes by setting $\boldsymbol{\omega}=\mathbf{0}$. Because this study considers only revolute joints, a zero axis should be interpreted solely as a placeholder rather than as an indicator of prismatic motion. For missing positions, we preserve spatial context by imputing $\mathbf{q}$ with the mean of the existing joints in the same group; when a group is entirely absent, $\mathbf{q}$ is inherited from the nearest parent group along the kinematic hierarchy. This scheme retains positional structure through $\mathbf{q}$ while keeping the axis channel uninformative wherever data are missing. \looseness = -1

While the $(\boldsymbol{\omega},\mathbf{q})$ representation is interpretable and globally consistent, the resulting design vector remains high-dimensional, making direct search inefficient and brittle. In the next subsections, we therefore learn a low-dimensional \emph{latent design space} from these screw-theoretic features and impose isometric regularization to obtain an optimization-friendly embedding.

\subsection{Humanoid Model Data Curation}

We constructed our dataset using 30 diverse humanoid and bi-arm robot models~\cite{digit,stasse2017talos,menagerie2022github,cremer2016baxter,ergocub,metta2010icub,alex,example-robot-data,radford2015valkyrie,pitzer2012pr2remote,kojima2015jaxon,Okugawa2015jvrc,lapeyre2014poppy,diftler2011r2,nextage,sciurus,allgeuer2016igusOP,asfour2013armar4,asfour2019armar6,pepper,kaneko2011hrp4,bruce,x1,rby1}, including representative examples such as Talos~\cite{stasse2017talos}, Baxter~\cite{cremer2016baxter}, Valkyrie~\cite{radford2015valkyrie}, and PR2~\cite{pitzer2012pr2remote}, selected to cover a wide range of sizes, joint configurations, and application domains. The complete list of 30 robots used in the dataset is visualized in Figure~\ref{fig:discussion}. Each robot was simulated in MuJoCo~\cite{todorov2012mujoco} and posed in an A-pose (or the closest feasible configuration) to standardize joint orientations, particularly at the elbows. For robots without existing MJCF files, URDF models were converted to MJCF. Global joint positions and axes were recorded in the world coordinate frame.

Joints were categorized into $14$ anatomical groups: torso and neck (no side), plus shoulder girdle, shoulder, upper arm, elbow, forearm, and wrist (left/right counterparts). Virtual joints were inserted only for missing wrists, as they serve as end-effectors and are necessary for maintaining structural consistency.\looseness=-1

For each group, we determined the maximum joint count across all robots and padded missing entries accordingly, imputing positions with the group mean when the group was partially present and inheriting from the nearest parent when it was entirely absent. The axis channel of padded entries followed the orientationless placeholder rule defined in our screw-theoretic representation, keeping the axis information intentionally uninformative wherever data were missing.
\looseness=-1

Positions were normalized by the shoulder-to-base distance to ensure size invariance. To reduce redundancy and enforce symmetry, only right-side joints (excluding torso and neck) were retained in the data representation. This compact representation maintains morphological completeness through mirror reconstruction during optimization. The final representation for each robot was a $(20,6)$ array—20 joints with $(\boldsymbol{\omega},\mathbf{q})$ screw parameters—flattened into a 120-dimensional vector for downstream learning.

% \begin{figure}[!t]                % !t: 가능한 Top 선호
%   \centering
%   \includegraphics[width=\columnwidth]{figures/robot_images.pdf} % 컬럼폭 사용
%   \caption{30 robots displayed in latent space, each on its own encoded latent vector}
%   \label{fig:top-right}
% \end{figure}

\subsection{Design manifold Learning}
%%% 3. Design manifold learning
%% ??? 
%% 

We trained an autoencoder, an encoder $f_\theta$ and decoder $g_\theta$ modeled with neural networks, with isometric regularization~\cite{lee2022regularized} on the 120-dimensional screw-based vectors from our curated dataset. The network comprises two hidden layers (64 and 32 units) and a latent dimension of 2 (with additional tests at 3), using Tanh activations and L2 reconstruction loss. Isometric regularization encourages the local Jacobian to approximate the identity matrix, preserving geometric continuity and reducing curvature in the latent space -- especially beneficial for small datasets to improve stability and generalization.

\subsection{Optimization within the Manifold}
Given the learned latent design space, we search latent vectors $\mathbf{z}$ whose decoded kinematic structures $g_\theta(\mathbf{z})$ can best reproduce the source motion $M_{\text{src}}$. Specifically, we minimize
\begin{IEEEeqnarray}{rCl}
\mathcal{L}_{\text{total}}(\mathbf{z}; M_{\text{src}})
&=& {\cal L}_{\rm PA-MPJPE} \Big(M_{\text{src}},\, \mathcal{R}(g_\theta(\mathbf{z}); M_{\text{src}})\Big) \nonumber\\
&& {}+ \lambda_{\text{joint}}\, N_{\text{tot}} \big(g_\theta(\mathbf{z})\big).
\end{IEEEeqnarray}
where $\mathcal{R}$ retargets $M_{\text{src}}$ onto the decoded structure, and the PA-MPJPE~\cite{dabral2018learning} loss aligns motions via a Procrustes transform and then computes mean per-joint position error. 
$N_{\text{tot}}(\cdot)$ counts the \emph{total} number of active joints in the mirrored full upper body; a joint $j$ is considered active if $\|\boldsymbol{\omega}_j\|_2\geq\varepsilon$, which ignores padded placeholders. 
This penalty discourages gratuitous articulation and biases the search toward parsimonious, task-specific designs.

We optimize $\mathcal{L}_{\text{total}}$ over $\mathbf{z}$ using VOO, a gradient-free method effective in low-dimensional continuous spaces. 
Because our representation stores only right-side joints (plus torso and neck), we mirror the decoded right side across the sagittal plane to obtain the full upper body before evaluation and count $N_{\text{tot}}$.\looseness=-1

\section{Experiments}
 % We evaluate motion‑to‑structure design optimization on three upper‑body CMU Mocap sequences (\textit{Wave Hello}, \textit{Chicken Dance}, \textit{Swimming}). We compare representations and learning variants—\textbf{Screw–120D (direct)}, \textbf{DH+IsoAE}, \textbf{Screw+AE} (no isometry), and \textbf{Screw+IsoAE (ours)}—and ablate the latent dimensionality (2D vs.\ 3D). Candidate structures are decoded from $(w,q)$ features, mirrored to full upper body, and searched with Voronoi Optimistic Optimization (VOO) under a fixed budget with ten shared seeds and a frozen decoder ($n_{\text{init}}{=}16$ uniform starts $+$ $T{=}30$ iterations $\Rightarrow$ $46$ evaluations per run). The total objective is $J=\mathrm{PA\text{-}MPJPE}+\lambda_{\text{joint}}N_{\mathrm{tot}}$ with $\lambda_{\text{joint}}=3.5$ and an activation rule $\|w_j\|_2\ge\varepsilon$ ($\varepsilon=0.5$). \textbf{Experiment~1} reports best‑within‑budget scores per motion and jointly over all three; \textbf{Experiment~2} analyzes sample efficiency on \textit{Chicken Dance} via mean best‑so‑far curves with IQR/min–max bars and an oracle histogram. 
% \subsection{Overview}
% motion -> design 
In this section, we evaluate our proposed method on three upper-body CMU Mocap sequences through two primary experiments.
First, we demonstrate the effectiveness of our screw-theory representation within an AE latent space by comparing it against the traditional DH parameterization and direct optimization in the naive 120-dimensional space, as detailed in Section~\ref{sec:representation}. Second, we validate that applying isometric regularization improves designs and that a compact, two-dimensional latent space is sufficient for effective optimization, as shown in Section~\ref{sec:ablation}.

\subsection{Experimental Setup}

\setupitem{Motions.}{We use three CMU Mocap sequences as $M_\text{src}$: \textit{141\_16 Wave Hello} (arm‑dominant; negligible torso), \textit{18\_15 Chicken Dance} (moderate torso yaw), and \textit{79\_02 Swimming} (torso yaw+pitch with dynamic arms)~\cite{cmumocap2007}.}
\setupitem{Search space \& optimizer.}{Candidate structures are decoded as $g_\theta(\mathbf{z})$ from $(\boldsymbol{\omega},\mathbf{q})$ features and mirrored to obtain the full upper body; for the retargeter $\mathcal{R}$, we used optimization-based retargeting method~\cite{jeong2025robust}. We search over $\mathbf{z}$ with VOO under a fixed evaluation budget shared across methods.}
\setupitem{Objective \& hyperparameters.}{We minimize PA-MPJPE~\cite{dabral2018learning}  penalized with number of joint, $J=\mathrm{PA\text{-}MPJPE}+\lambda_{\text{joint}}N_{\mathrm{tot}}$, counting a joint active if $\|\boldsymbol{\omega}_j\|_2\ge\varepsilon$; we set $\lambda_{\text{joint}}=3.5$ and $\varepsilon=0.5$. For DH baselines, near‑zero links are clamped to zero when $|d|\le 0.01$ or $|a|\le 0.01$.}
% \setupitem{Baselines and variants.}{\je{We compare our proposed method, against several baselines to validate its core components. We use \textbf{Baseline}, which is screw-theory based representation of 120 dimensions, to show the benefit of a learning manifold, and \textbf{DH}~\cite{} to prove the superiority of the screw-theory representation. Furthermore, we compare the component of the use of isometric regularizator, }}
\setupitem{Protocol.}{All methods use the same evaluation budget and ten shared seeds; the decoder $g_\theta$ is frozen during optimization. VOO searches the latent box $\mathbf{z}\in[-15,15]^d$ ($d{=}2$ for main results; $d{=}3$ for the ablation) with $n_{\text{init}}{=}16$ uniformly drawn starts and $T{=}30$ subsequent iterations (one evaluation each), i.e., $n_{\text{eval}}{=}n_{\text{init}}+T=46$ evaluations per run. Unless otherwise noted, we also report the total objective averaged over the three motions together with its two components (PA‑MPJPE and $N_{\mathrm{tot}}$).}

% \paragraph{Experiment 1}

% \paragraph{Experiment 2: VOO on \textit{Chicken Dance}}
% \begin{figure}                % !t: 가능한 Top 선호
%   \centering
%   \includegraphics[width=\columnwidth]{figures/experiment2.png} % 컬럼폭 사용
%   \caption{Error bars: IQR (thick) and min–max (thin). Left inset: oracle distribution with dashed mean $\mu$.}
%   \label{fig:top-right}
% \end{figure}
% On \textit{18\_15 Chicken Dance}, \textbf{2D IsoAE} consistently achieves the lowest mean best‑so‑far objective across iterations: it descends faster over the first 1–5 evaluations, reaches the lowest plateau by step 25, and shows tighter dispersion than \textbf{3D IsoAE}, indicating that a two‑dimensional latent design space suffices for this task. Compared with \textbf{2D AE} (no isometry), \textbf{2D IsoAE} remains uniformly better across the entire budget, evidencing the benefit of the isometric regularizer in shaping a well‑behaved latent geometry that improves sample efficiency and robustness. In contrast, the \textbf{Screw–120D (direct)} baseline starts substantially higher, improves slowly, and stays near or above the oracle mean with wide IQR/min–max bands, highlighting the difficulty of high‑dimensional search without representation learning.

\subsection{Representation and Manifold of Robot Structure} \label{sec:representation}

We present a quantitative comparison of our method in Table~\ref{tab:results} to show the effectiveness of screw-based representation and manifold learning. We report the best‑within‑budget total objective (mean$\pm$std over 10 seeds) \textit{per} motion and also for the joint objective over \textit{set} of all three motions (sum of the three retargeting terms plus one joint‑count penalty). To evaluate our approach, we compare our method against two baselines: Denavit–Hartenberg parameters (DH)~\cite{DenavitHartenberg1955,HartenbergDenavit1964}, which uses a learned manifold of traditional DH parameters, and a direct screw-theory based representation (baseline), which performs direct optimization on the high-dimensional parameter space. 
We implemented standard DH parameterization, augmenting it with an axis offset to account for the humanoid upper-body topology—from the torso to the neck and shoulders—and for robots whose wrists terminate as links without an explicit joint, we introduced separate tool center points (TCP) for each wrist in addition to the 14 anatomical joints.

% \paragraph{DH vs Screw-theory}

% Table~\ref{tab:results} reports the proposed method is consistently best: $97.50\pm3.68$ (Wave Hello), $100.65\pm5.07$ (Chicken Dance), $101.30\pm3.60$ (Swimming), and $178.46\pm10.89$ (all‑motions). 
Compared with DH~\cite{DenavitHartenberg1955,HartenbergDenavit1964} parameter-based representation, the reductions are 39.9\% (Wave), 32.4\% (Dance), 32.9\% (Swimming) with a single‑motion average of 35.0\%, and 43.5\% on the all‑motions objective.  The superior result of the screw-theoretic representation for optimization indicates that the screw‑theoretic representation with isometric regularization provides both superior accuracy and stability, especially when a single structure must accommodate heterogeneous motions.
% The mean gaps are large compared to the reported standard deviations (e.g., Wave: $54.7$ absolute points vs.\ $13.95/3.68$ stds for direct/ours), and ours exhibits lower across‑seed variability (single‑motion CV $3.6$–$5.0\%$ vs.\ $6.7$–$10.0\%$ for direct and $10.8$–$13.9\%$ for DH). 
% DH+IsoAE is comparable to or slightly better than direct on Chicken and Swimming but worse on Wave, and it underperforms direct on the all‑motions objective ($315.88\pm27.39$ vs.\ $274.44\pm33.94$), indicating that the screw‑theoretic representation with isometric regularization provides both superior accuracy and stability, especially when a single structure must accommodate heterogeneous motions.

% \paragraph{Manifold Learning}
% Relative to \textbf{Screw–120D (direct)}, this yields reductions of \textbf{36.0\%}, \textbf{34.2\%}, and \textbf{34.6\%} on the three single motions (avg.\ \textbf{34.9\%}), and \textbf{35.0\%} on the all‑motions objective. 
By comparing our full method to baseline, which uses the same screw-theory parameters but optimizes in the full 120-dimensional space without manifold learning, we observe massive and consistent performance gains. This approach yields objective score reductions of 36.0\%, 34.2\%, and 34.6\% on the individual motions, and 35.0\% on the joint all-motions task. This large margin highlights the difficulty of navigating high-dimensional parameter spaces, as learning a structured manifold of valid robot designs makes the optimization problem more tractable.

% Screw–120D (direct)
\newcommand{\sdWHmean}{152.2}\newcommand{\sdWHstd}{13.9}
\newcommand{\sdCDmean}{153.0}\newcommand{\sdCDstd}{10.2}
\newcommand{\sdSWmean}{154.8}\newcommand{\sdSWstd}{15.4}
\newcommand{\sdAmean}{274.4}\newcommand{\sdAstd}{33.9}

% DH+IsoAE
\newcommand{\dhWHmean}{162.1}\newcommand{\dhWHstd}{17.4}
\newcommand{\dhCDmean}{148.8}\newcommand{\dhCDstd}{17.3}
\newcommand{\dhSWmean}{150.9}\newcommand{\dhSWstd}{20.9}
\newcommand{\dhAmean}{315.8}\newcommand{\dhAstd}{27.3}

% Screw+IsoAE (ours)
\newcommand{\scWHmean}{97.5}\newcommand{\scWHstd}{3.6}
\newcommand{\scCDmean}{100.6}\newcommand{\scCDstd}{5.0}
\newcommand{\scSWmean}{101.3}\newcommand{\scSWstd}{3.6}
\newcommand{\scAmean}{178.4}\newcommand{\scAstd}{10.8}
\begin{table}[!t]
  \centering
  \caption{Best-within-budget total objective on three motions (mean$\pm$std over 10 seeds; lower is better). Objective is $\mathrm{PA\text{-}MPJPE}+\lambda_{\text{joint}}N_{\mathrm{tot}}$ with $\lambda_{\text{joint}}=3.5$ and $\varepsilon=0.5$.}
  \label{tab:results}
  \resizebox{0.99\linewidth}{!}{%
  \begin{tabular}{l|ccc|c}
    \toprule
    \multirow{2}{*}{Rep.} & \multicolumn{3}{c|}{Motion($\downarrow$)} & \multirow{2}{*}{Motion Set($\downarrow$)} \\
    & Wave & Dance & Swimming & \\
    \midrule
    Baseline & \ms{\sdWHmean}{\sdWHstd} & \ms{\sdCDmean}{\sdCDstd} & \ms{\sdSWmean}{\sdSWstd} & \ms{\sdAmean}{\sdAstd} \\
    DH~\cite{DenavitHartenberg1955}  & \ms{\dhWHmean}{\dhWHstd} & \ms{\dhCDmean}{\dhCDstd} & \ms{\dhSWmean}{\dhSWstd} & \ms{\dhAmean}{\dhAstd} \\
    \textbf{Ours}   & \bfseries\ms{\scWHmean}{\scWHstd} & \bfseries\ms{\scCDmean}{\scCDstd} & \bfseries\ms{\scSWmean}{\scSWstd} & \bfseries\ms{\scAmean}{\scAstd} \\
    \bottomrule
  \end{tabular}}
\end{table}

\subsection{Role of Isometric Regularization and Latent Dimension} \label{sec:ablation}
\begin{figure}[t]
  \centering
  \includegraphics[width=\columnwidth]{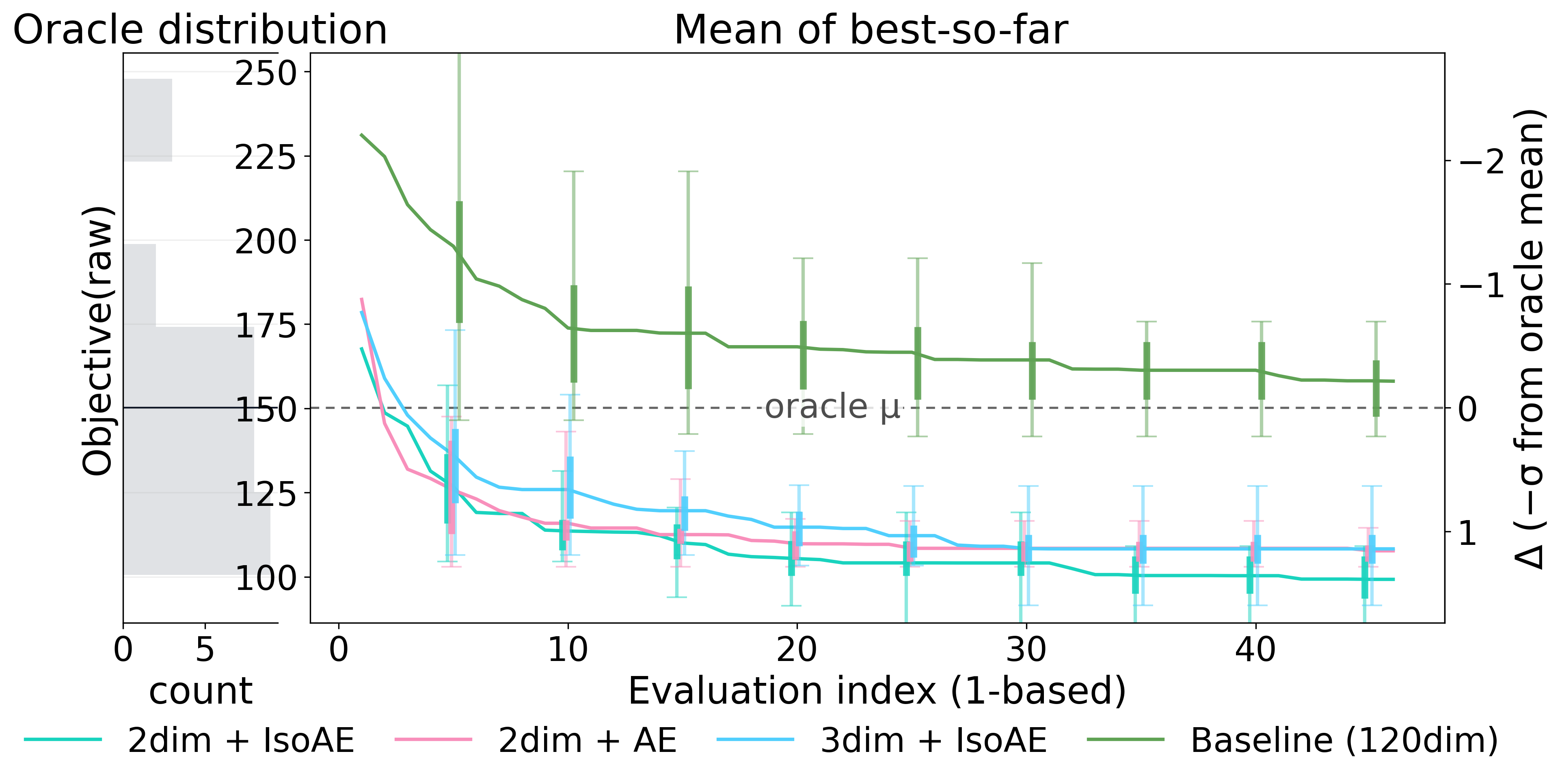}
  \caption{Mean best‑so‑far objective across ten shared seeds(lower is better). The error bars represent the interquartile range (thick) and min–max (thin). Left inset: oracle distribution with dashed mean $\mu$ on motion \textit{chicken dance}.}
  \label{fig:exp2_chicken}
  \vspace{-5mm}
\end{figure}
\begin{figure*}[!t]                % !t: 가능한 Top 선호
  \centering
  \includegraphics[width=\linewidth]{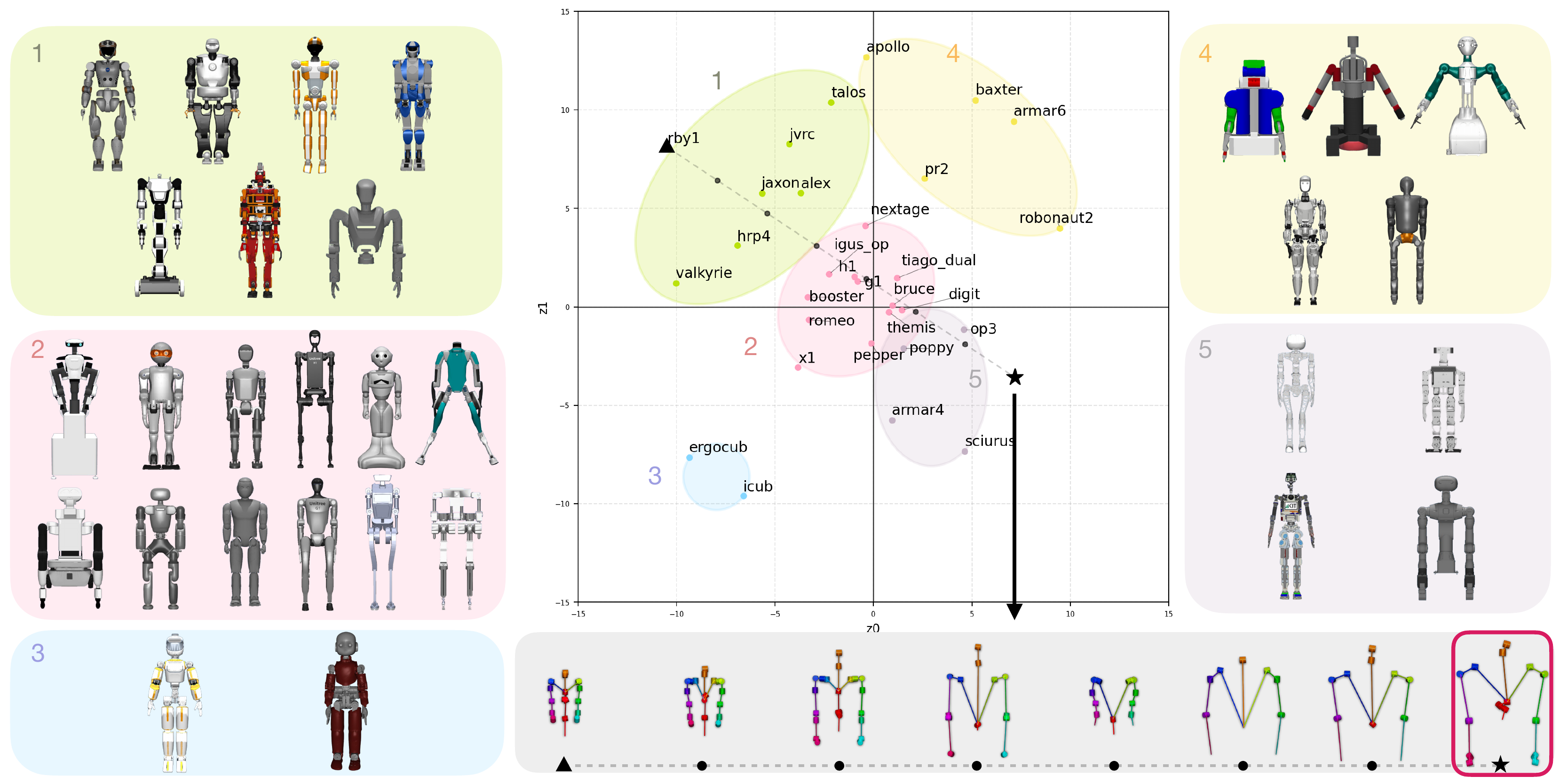} % 컬럼폭 사용
  \caption{Latent map of 30 robots (2D IsoAE). Colors show $k$‑means clusters ($k{=}5$) with representative thumbnails, which demonstrate that the robots are semantically grouped based on their morphological similarity (e.g., full-body humanoids, dual-arm manipulators). The bottom strip interpolates eight points from \texttt{rby1} to a starred design (sufficient for \textit{Swimming}); joint birth/death is visible as markers appear/disappear under the activation rule.}
  \label{tab:discussion}
  \label{fig:discussion}
  \vspace{-5mm}
\end{figure*}

Figure \ref{fig:exp2_chicken} shows the mean best‑so‑far objective across ten shared seeds under a fixed evaluation budget on \textit{18\_15 chicken dance} motion compared with the use of the isometric regularizer (2D AE~\cite{HintonSalakhutdinov2006,GoodfellowBengioCourville2016}) and latent dimension size (2D IsoAE and 3D IsoAE). 
2D IsoAE (Ours) achieves the lowest mean best‑so‑far objective over the entire 46‑evaluation budget. It drops sharply in the first 5–10 evaluations, falls below the oracle mean $\mu$ early, and then continues to improve monotonically before stabilizing around steps 30–35; its dispersion is tighter than 3D IsoAE, indicating that a two‑dimensional latent design space suffices. Compared with 2D AE (no isometry), 2D IsoAE remains uniformly better across all steps, evidencing the benefit of the isometric regularizer. In contrast, baseline (screw representation with 120 dimensions) starts higher, improves slowly, and stays near or above $\mu$ with wide interquartile range (IQR)/min–max bands, highlighting the difficulty of high‑dimensional search without representation learning.

Beyond the ranking, three properties stand out. \emph{Sample efficiency}: within the first ten evaluations, 2D IsoAE already sits well below $\mu$ and separates on the normalized right axis, while 3D IsoAE and 2D AE lag and the 120D baseline hovers around $\mu$. \emph{Robustness}: the IQR of 2D IsoAE tightens quickly and its min–max whiskers shrink after roughly 20 evaluations; by the final budget, the IQR bands of 2D IsoAE have little overlap with those of 3D IsoAE or 2D AE, indicating lower seed sensitivity. \emph{Dimension effect}: under a fixed budget, increasing latent dimensionality dilutes VOO’s sampling density (Voronoi cells effectively shrink in higher $d$), which hurts exploitation—consistent with the slower progress of 3D IsoAE and the 120D direct baseline.
\begin{figure}[!t]                % !t: 가능한 Top 선호
  \includegraphics[width=\columnwidth ]{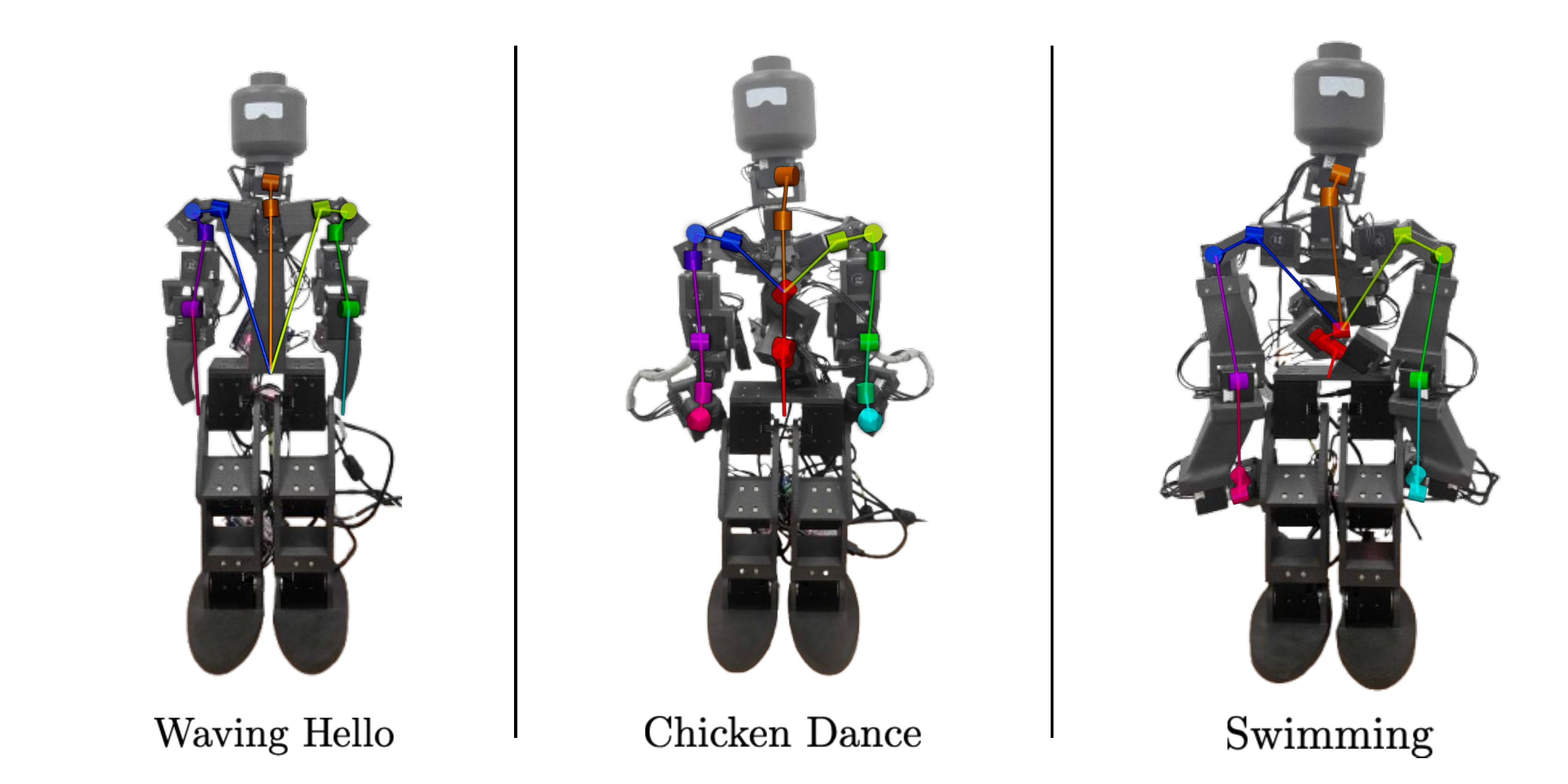} % 컬럼폭 사용
  \caption{Hardware prototypes generated by our design framework under manufacturing constraints. Each prototype is optimized for a specific task (`Waving Hello', `Chicken Dance', `Swimming'), resulting in different kinematic structures.}
  \label{fig:kinematic_structure}
  \vspace{-5mm}
\end{figure}
\begin{figure*}[!t]                % !t: 가능한 Top 선호
  \centering
  \includegraphics[width=0.9\linewidth]{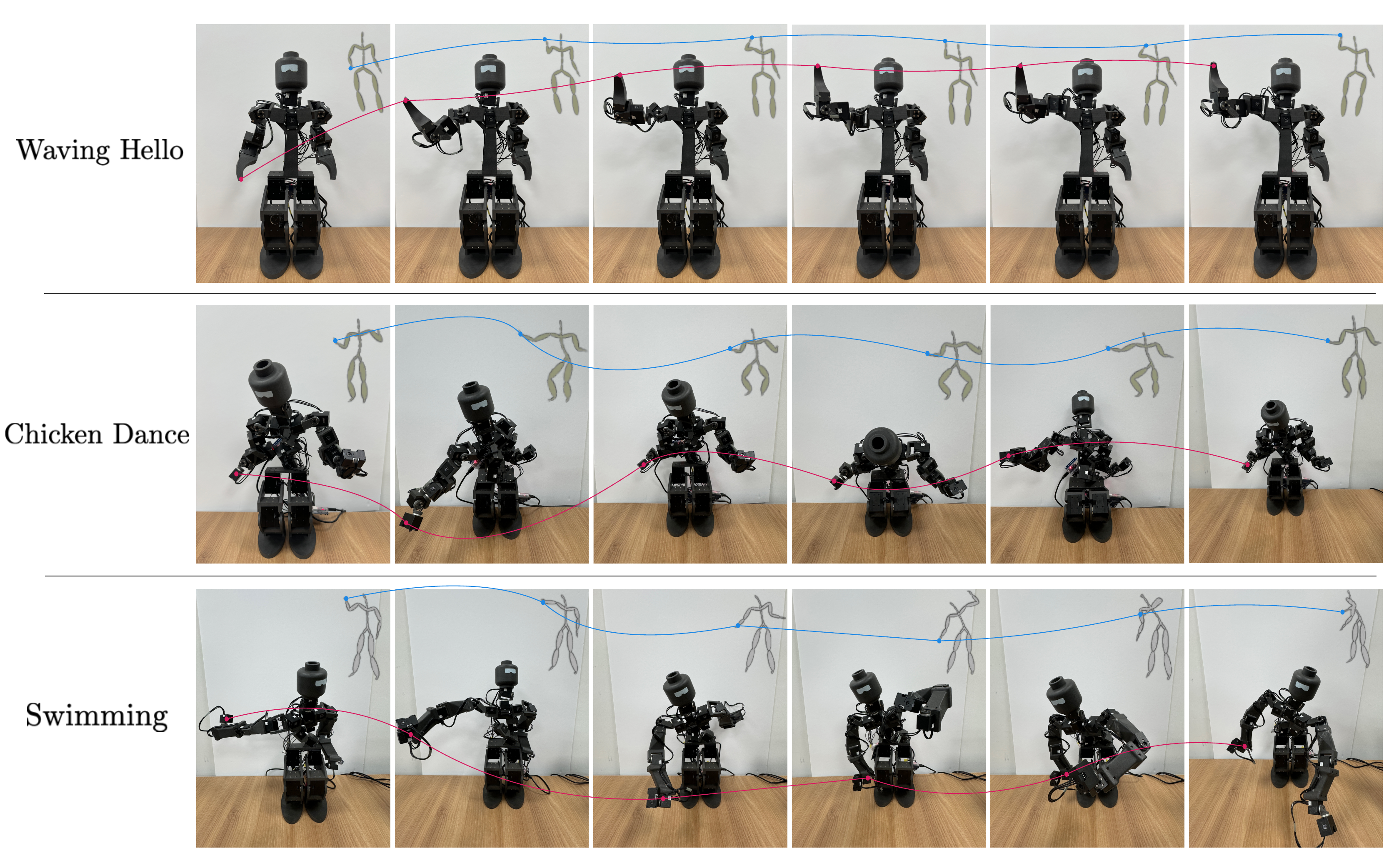} % 컬럼폭 사용
  \caption{Demonstration of the three motion-optimized robots from Fig.~\ref{fig:kinematic_structure} successfully executing Waving Hello, Chicken Dance, and Swimming via motion retargeting. For each robot snapshot, the human figure (upper right) shows the key pose from the source human motion (BVH file) used as input, while the photo shows the retargeted execution on the robot's unique kinematics. The red and blue lines visualize the 3D end-effector trajectories of robot and source human motion.}
  \label{fig:hardware}
  \vspace{-5mm}
\end{figure*}
\subsection{Discussion: structure of the learned design space}
Beyond its quantitative performance, our model learns a meaningful latent space for robot design. This is visualized in Figure~\ref{tab:discussion}, which shows how the manifold groups similar structures into distinct $k$-means clusters ($k{=}5$) with representative thumbnails. The clusters align with intuitive morphology families: a ``large full‑body humanoids'' group (e.g., Talos~\cite{stasse2017talos}, JAXON~\cite{kojima2015jaxon}, rby1~\cite{rby1}, etc.) occupies the upper‑left region; dual‑arm mobile‑manipulators and upper‑torso platforms (Baxter~\cite{cremer2016baxter}, PR2~\cite{pitzer2012pr2remote}, etc.) reside in the upper‑right; compact social/service humanoids (Pepper, OP3, H1, etc.) concentrate near the origin; the iCub family (iCub~\cite{metta2010icub}, ergoCub~\cite{ergocub}) forms a distinct basin in the lower‑left; and a lighter research‑platform cluster (e.g., ARMAR‑4~\cite{asfour2013armar4}, Sciurus~\cite{sciurus}) appears in the lower‑right. These groupings suggest that the isometric latent space preserves semantically meaningful kinematic traits (torso involvement, shoulder‑girdle complexity, anthropomorphic proportion), providing a navigable atlas for design search.

Interpolating through the latent space reveals both smooth morphing of link layouts and discrete changes like joint birth/death, showing that our decoder accommodates variable joint counts. Notably, optimal solutions for certain motions (starred) can lie on cluster boundaries, suggesting they blend features from different design families. Overall, the space provides: (i) coherent clustering, (ii) smooth and discrete kinematic changes, and (iii) actionable starting points for future optimization.

% The interpolation strip (bottom) traces eight evenly spaced points from rby1 to the starred location (a sub‑optimal yet sufficient solution for \textit{Swimming}); along this path we observe smooth morphing in link layout \emph{and} discrete reconfiguration via joint birth/death (activation threshold on $\|w_j\|$), indicating that the decoder supports continuous transitions while naturally accommodating variable joint counts. Notably, the starred region lies near the boundary between clusters, suggesting that motion‑specific optima can blend features across morphology families (e.g., redistributing torso vs.\ shoulder articulation for \textit{Swimming}). Overall, the map offers (i) semantically coherent clustering, (ii) continuous–and occasionally discrete–kinematic changes under interpolation, and (iii) actionable waypoints for initializing or constraining VOO in downstream tasks.

\subsection{Hardware}

To empirically validate the pipeline, we fabricated a sufficiently good hardware prototype under manufacturing and schedule constraints. All kinematic structures were denormalized by applying a single scale factor such that, relative to a 170mm lower‑body reference, the base‑to‑neck distance equals 124mm. Assembly was performed in the same A‑pose world frame used throughout the dataset. Joint axes and reference positions were nominally preserved; where unavoidable to accommodate packaging and assembly, we permitted local position adjustments up to 10mm (direction‑agnostic). Left–right geometry was obtained by geometric mirroring about the sagittal plane, and all subsequent retargeting and evaluation used the as‑built CAD/MJCF model. On the as‑built platform, we executed the retargeted joint trajectories for all three source motions—\textit{Wave Hello}, \textit{Chicken Dance}, and \textit{Swimming}—using the same world frame and activation rule as in simulation. Without any architectural changes to the pipeline, the as-built upper-body humanoids smoothly tracked the commanded poses (see Fig.~\ref{fig:hardware}).

\subsection{Limitations}
While the retargeting‑based objective promotes task fidelity and parsimony, our decoder and search do not yet enforce hard manufacturability constraints. In particular, we do not impose (i) kinematic feasibility such as discrete joint types and range limits, (ii) geometric validity including self‑collision avoidance, mesh/packaging/clearance constraints, and (iii) dynamic plausibility with actuation limits, torque/energy budgets, and contact stability. Future work will integrate these constraints via a structure‑aware decoder (e.g., feasibility‑projected decoding), lightweight feasibility filters (reject/repair with collision and range checks), and simulator‑backed penalties (inverse dynamics, torque limits), moving toward multi‑objective design and real‑hardware validation. \looseness=-1

\section{Conclusion}
% 논문의 contributions 정리 및 향후 연구 방향 언급

We presented a motion-to-structure pipeline for upper-body humanoid design optimization. The approach hinges on three components: (i) a screw-theoretic joint representation $(\boldsymbol{\omega},\mathbf{q})$ that encodes global joint axes and positions and enables consistent cross-morphology reasoning; (ii) a compact latent design space learned with an isometric-regularized autoencoder to provide a smooth, optimization-friendly manifold; and (iii) gradient-free search in this manifold using VOO, driven by a \emph{total objective} that combines PA-MPJPE after retargeting with a mild joint-count penalty. Together, these elements turn human motion priors into actionable guidance for selecting kinematic structures, achieving ${\sim}34\%$ lower average objective than direct search and ${\sim}39\%$ than DH+IsoAE across three motions. 
\looseness = -1

% Experiments on three CMU Mocap sequences (\textit{Wave Hello}, \textit{Chicken Dance}, \textit{Swimming}) show that the proposed \textbf{Screw+IsoAE} consistently achieves lower total objective and faster convergence than both \textbf{DH+IsoAE} and direct search in the original 120-D screw space (\textbf{Screw-120D}). Moreover, a 2-D latent manifold is sufficient for efficient search—matching or surpassing a 3-D counterpart—while isometric regularization improves both stability and final performance. The joint-count penalty steers the optimizer toward parsimonious, task-specific designs (e.g., minimal torso articulation for \textit{Wave Hello} versus increased yaw/pitch capacity for \textit{Swimming}), supporting our claim that distinct motions admit distinct optimal upper-body kinematic structures.

% \paragraph{Limitations and future work.}
% While the retargeting-based objective encourages feasible designs, the decoder does not explicitly guarantee physical validity. Future work will integrate hard kinematic/dynamic constraints (e.g., joint limits, collisions, torque/energy costs) and extend the objective to multi-objective settings. We also plan to explore differentiable or learned retargeters, richer design variables (e.g., link mass/inertia, actuation), whole-body and hand augmentation, and real-hardware validation. These directions aim to broaden applicability while preserving the sample efficiency and interpretability of the proposed motion-driven design pipeline.

% \clearpage
\bibliographystyle{IEEEtran}
\bibliography{references2}
\appendix
\renewcommand{\thesection}{\Alph{section}}
\renewcommand{\thesubsection}{\Alph{section}.\arabic{subsection}}

\vspace{1em}
\setcounter{section}{0}
\refstepcounter{section}\label{app:robot-dataset}
\noindent{\large\textbf{Appendix A. Robot Dataset}}
%% ============================================================
% \section{Robot Dataset}
% \label{app:robot-dataset}

%% --- A.1 Robot List ---
\subsection{Robot List}

We use 30 diverse robots collected from publicly available MJCF/URDF assets.
Table~\ref{tab:robot-list} lists all robots included in the dataset.
Each robot is classified as either Humanoid (bipedal, with articulated legs) or Non-bipedal (no legs, mounted on a fixed pedestal, wheeled base, or similar platform).
Only the upper-body kinematics are used for all robots, and the legs of bipedal robots are kept in a fixed pose during data extraction.

\begin{table}[H]
\centering
\caption{Robots included in the dataset.}
\label{tab:robot-list}
\small
\setlength{\tabcolsep}{4.5pt}
\renewcommand{\arraystretch}{1.08}
\resizebox{\columnwidth}{!}{%
\begin{tabular}{clc@{\hspace{10pt}}clc}
\toprule
\# & Robot & Type & \# & Robot & Type \\
\midrule
1  & Alex~\cite{alex}              & Non-bipedal & 16 & JVRC~\cite{Okugawa2015jvrc}       & Humanoid     \\
2  & Apollo~\cite{menagerie2022github}  & Humanoid    & 17 & Nextage~\cite{nextage}        & Non-bipedal  \\
3  & ARMAR-4~\cite{asfour2013armar4}   & Humanoid    & 18 & OP3~\cite{menagerie2022github}    & Humanoid     \\
4  & ARMAR-6~\cite{asfour2019armar6}   & Non-bipedal & 19 & Pepper~\cite{pepper}          & Non-bipedal  \\
5  & Baxter~\cite{cremer2016baxter}    & Non-bipedal & 20 & Poppy~\cite{lapeyre2014poppy}     & Humanoid     \\
6  & Booster~\cite{menagerie2022github} & Humanoid    & 21 & PR2~\cite{pitzer2012pr2remote}    & Non-bipedal  \\
7  & Bruce~\cite{bruce}            & Humanoid    & 22 & RBY1~\cite{rby1}              & Non-bipedal  \\
8  & Digit~\cite{digit}            & Humanoid    & 23 & Robonaut2~\cite{diftler2011r2}    & Humanoid     \\
9  & ergoCub~\cite{ergocub}         & Humanoid    & 24 & Romeo~\cite{example-robot-data}   & Humanoid     \\
10 & G1~\cite{menagerie2022github}     & Humanoid    & 25 & Sciurus~\cite{sciurus}         & Non-bipedal  \\
11 & H1~\cite{menagerie2022github}     & Humanoid    & 26 & TALOS~\cite{stasse2017talos}      & Humanoid     \\
12 & HRP-4~\cite{kaneko2011hrp4}      & Humanoid    & 27 & Themis~\cite{bruce}           & Humanoid     \\
13 & iCub~\cite{metta2010icub}       & Humanoid    & 28 & TIAGo Dual~\cite{menagerie2022github} & Non-bipedal  \\
14 & IGUS OP~\cite{allgeuer2016igusOP}  & Humanoid    & 29 & Valkyrie~\cite{radford2015valkyrie} & Humanoid     \\
15 & JAXON~\cite{kojima2015jaxon}     & Humanoid    & 30 & X1~\cite{x1}                & Humanoid     \\
\bottomrule
\end{tabular}}
\end{table}
%% --- A.2 Joint Classification Taxonomy ---
\subsection{Joint Classification Taxonomy}

Each robot's upper-body joints are classified into 14 anatomical groups for the screw representation.
The groups are listed in Table~\ref{tab:joint-groups}.
For the DH representation, two additional groups, left and right Tool Center Point (TCP) are appended to represent virtual end-effectors, yielding 16 groups in total.
When exploiting bilateral symmetry, only eight groups are used (torso, neck, and the six right-side groups), with the left side reconstructed via mirroring.

\begin{table}[!t]
\centering
\caption{Joint classification groups (screw representation, 14 groups).}
\label{tab:joint-groups}
\resizebox{0.99\linewidth}{!}{%
\begin{tabular}{l|l}
\toprule
Group & Description \\
\midrule
Torso                        & Trunk joints originating from the base link \\
Neck                         & Neck joints following the torso chain \\
Left / Right Shoulder Girdle & Clavicle and scapula region \\
Left / Right Shoulder        & Primary shoulder joint \\
Left / Right Upper Arm       & Upper arm rotation \\
Left / Right Elbow           & Elbow flexion / extension \\
Left / Right Forearm         & Forearm pronation / supination \\
Left / Right Wrist           & Wrist articulation \\
\bottomrule
\end{tabular}}
\end{table}

%% --- A.3 Max Joints per Group ---
The maximum number of joints per group across all 30 robots
is summarized in Table~\ref{tab:max-joints}.
With six screw parameters per joint, the 14-group layout
yields $30 \times 6 = 180$ features.
For the DH representation, two virtual TCP groups are
appended---one per arm---to anchor the end-effector frame,
yielding 32 joint slots and $32 \times 5 = 160$ features.

\begin{table}[h]
\centering
\caption{Maximum joint count per anatomical group.
For the DH representation, one TCP group per arm is appended
(1 slot each), giving 16 groups and 32 slots in total.}
\label{tab:max-joints}
\small
\begin{tabular}{lc|lc}
\toprule
Group & Max & Group & Max \\
\midrule
Torso              & 6 & R. Shoulder Girdle & 1 \\
Neck               & 4 & R. Shoulder        & 3 \\
L. Shoulder Girdle & 1 & R. Upper Arm       & 1 \\
L. Shoulder        & 3 & R. Elbow           & 1 \\
L. Upper Arm       & 1 & R. Forearm         & 1 \\
L. Elbow           & 1 & R. Wrist           & 3 \\
L. Forearm         & 1 & (R. TCP)           & 1 \\
L. Wrist           & 3 & & \\
(L. TCP)           & 1 & & \\
\midrule
\multicolumn{2}{l}{Screw total (14 groups)} & \multicolumn{2}{l}{30 slots} \\
\multicolumn{2}{l}{DH total (16 groups)}    & \multicolumn{2}{l}{32 slots} \\
\bottomrule
\end{tabular}
\end{table}
% Across the 30 robots, the maximum number of joints per group is
% $[\text{torso}\colon6,\;\text{neck}\colon4,\;\text{left}\_\text{shoulder}\_\text{girdle}\colon1,\;\text{left}\_\text{shoulder}\colon3,\;\text{left}\_\text{upperarm}\colon1,\;\text{left}\_\text{elbow}\colon1,\;\text{left}\_\text{forearm}\colon1,\;\text{left}\_\text{wrist}\colon3,\;\text{right}\_\text{shoulder}\_\text{girdle}\colon1,\;\text{right}\_\text{shoulder}\colon3,\;\text{right}\_\text{upperarm}\colon1,\;\text{right}\_\text{wrist}\colon1,\;\text{right}\_\text{forearm}\colon1,\;\text{right}\_\text{wrist}\colon3]$,
% total 30 joint slots.
% With six screw parameters per joint the full feature vector has 180 dimensions;
% with five DH parameters per joint it has 150 dimensions.

Because we exploit bilateral symmetry (Appendix~\ref{app:kinematic-repr}),
only the torso, neck, and right-side arm groups are retained for training,
discarding the left-side groups whose parameters are reconstructed by mirroring.
This yields 20 joint slots:
torso\,(6) + neck\,(4) = 10 for the central groups,
and shoulder girdle\,(1) + shoulder\,(3) + upper arm\,(1) + elbow\,(1) + forearm\,(1) + wrist\,(3) = 10 for the right arm,
corresponding to $20 \times 6 = 120$ features for screw and $21 \times 5 = 105$ for DH
(the extra slot comes from the right-side TCP group).

%% ============================================================
\vspace{1em}

\setcounter{section}{1}
\refstepcounter{section}\label{app:kinematic-repr}
\noindent{\large\textbf{Appendix B. Kinematic Representations}}
% \section{Kinematic Representations}
% \label{app:kinematic-repr}

%% --- B.1 Screw Axis ---
\subsection{Screw Axis Extraction (Product of Exponentials)}

% Each revolute joint is represented by a six-dimensional screw coordinate:
% a three-dimensional unit rotation axis and a three-dimensional position of the joint anchor, both expressed in the world frame.
% To extract these parameters, each robot is first placed in a predefined A-pose (a neutral standing posture with arms slightly abducted) using MuJoCo's forward kinematics.
Each revolute joint is represented by a six-dimensional screw coordinate:
a three-dimensional unit rotation axis and a three-dimensional position
of the joint anchor, both expressed in the world frame.
To extract these parameters, each robot is first placed in a canonical
A-pose (a neutral standing posture with arms slightly abducted) using
MuJoCo's forward kinematics.
The rotation axis is then read directly from the simulator's world-frame
joint axis, and the position is taken from the joint anchor computed in
that configuration.
No additional transformation or fitting is required, making the extraction
deterministic and exact for each MJCF model.
The extracted world-frame quantities are subsequently centered and scaled
as described in Appendix~\ref{app:normalization}.
% The rotation axis is then read directly from the simulator's world-frame joint axis,
% and the position is taken from the joint anchor computed in that configuration.
% No additional transformation or fitting is required, making the extraction deterministic and exact for each MJCF model.

%% --- B.2 Standard DH Parameters ---
\subsection{Standard DH Parameters Extraction}

As an alternative representation, each joint is described by five parameters:
the joint angle theta (rotation about the previous z-axis),
the link offset d (translation along the previous z-axis),
the link length a (distance along the common normal),
the twist angle alpha (rotation about the common normal),
and an additional axis offset tau.
The first four are the classical Denavit--Hartenberg parameters;
tau is an augmentation that captures the displacement of the joint anchor
along its own rotation axis relative to the DH-derived frame origin.
This extra parameter is necessary because standard DH conventions assume
that consecutive joint axes either intersect or share a well-defined common normal,
whereas real robot designs often place joints at arbitrary offsets along their axes.
Without tau, the reconstructed kinematic chain would accumulate positional errors
at every such offset.
All five parameters are computed numerically from consecutive joint-axis pairs
extracted from the MJCF model in the A-pose configuration.

%% --- B.3 Mirroring ---
\subsection{Mirroring (Symmetric to Full Body)}

To exploit bilateral symmetry, only the right-side joint groups (plus torso and neck) are encoded.
The left side is reconstructed by mirroring across the sagittal plane (the body's median plane, perpendicular to the y-axis).
For the screw representation, the y-components of both the rotation axis and the position vector are negated, reflecting all joint geometry through the y\,=\,0 plane.
For the DH representation, the joint angle theta and the twist angle alpha are negated while the link lengths d, a, and the axis offset tau are kept unchanged.
This is because reflecting a kinematic chain across the sagittal plane reverses the sense of rotation about each joint axis (hence theta flips sign) and reverses the twist between consecutive axes (hence alpha flips sign), while the translational offsets along and perpendicular to the axes remain the same.

%% ============================================================
% \section{Normalization}
% \label{app:normalization}
\vspace{1em}
\setcounter{section}{2}
\refstepcounter{section}\label{app:normalization}
\noindent{\large\textbf{Appendix C. Normalization}}

%% --- C.1 Position Normalization (Screw) ---
\subsection{Position Normalization (Screw)}

To make the screw representation comparable across robots of different scales, each robot's joint positions are first centered by subtracting the base-link position.
They are then divided by the robot's upper-body reference height, defined as the Euclidean distance between the average shoulder position and the base position.
This shoulder-to-base distance provides a consistent scale factor that is well-defined even for robots without a distinct neck joint.
Since the rotation-axis vectors are already unit-length, no further normalization is applied to them.

%% --- C.2 DH Normalization ---
\subsection{DH Normalization}

For the DH representation, the three length parameters (d, a, and tau) are scaled by the same shoulder-to-base distance used for the screw positions.
The two angle parameters (theta and alpha) remain unscaled.
In addition, the d-value of the first spine joint is rebased to account for the vertical offset of the base link from the world origin.

%% --- C.3 Missing Data Handling ---
\subsection{Missing Data Handling}

Not every robot possesses joints in all 14 groups.
A joint slot is considered missing when its rotation-axis norm falls below $10^{-2}$ during extraction.
For such slots, the axis entries are set to zero.
Position entries of missing slots are replaced by the mean position of the valid joints within the same group.
If the entire group is empty, a fallback hierarchy is used: for instance, a missing elbow position is substituted by the nearest available ancestor in the kinematic chain (upper arm, then shoulder, then torso, then base link).

During reconstruction from the autoencoder, a separate activation threshold of 0.5 is applied to the decoded axis vectors:
any axis whose norm falls below this value is zeroed out,
effectively deactivating joints that the decoder assigns negligible rotation to.
The extraction threshold ($10^{-2}$) identifies joints absent from the original robot model,
whereas the activation threshold (0.5) determines whether the decoder has assigned a meaningful axis to a reconstructed joint slot.

%% ============================================================
\vspace{1em}
\setcounter{section}{3}
\refstepcounter{section}\label{app:autoencoder}
\noindent{\large\textbf{Appendix D. Autoencoder Architecture and Training}}
% \section{Autoencoder Architecture and Training}
% \label{app:autoencoder}

%% --- D.1 Architecture ---
\subsection{Architecture}

The autoencoder uses a symmetric encoder--decoder structure.
The encoder maps the 120-dimensional input (20 joints times 6 screw parameters) through two hidden layers of size 64 and 32, respectively, down to the latent space of dimension $z$.
The decoder mirrors this structure, mapping from $z$ back to 120 dimensions through hidden layers of size 32 and 64.
All hidden layers use the Tanh activation function.
The latent dimension $z$ was swept from 2 to 3 across experiments.

\begin{table}[H]
\centering
\caption{Autoencoder architecture.}
\label{tab:ae-arch}
\resizebox{0.99\linewidth}{!}{%
\begin{tabular}{l|l}
\toprule
Component & Specification \\
\midrule
Input dimension    & 120 (20 joints $\times$ 6 screw parameters) \\
Hidden layers      & {[64, 32]} \\
Latent dimension   & 2 -- 3 (swept) \\
Activation         & Tanh \\
Encoder path       & 120 $\to$ 64 $\to$ 32 $\to$ $z$ \\
Decoder path       & $z$ $\to$ 32 $\to$ 64 $\to$ 120 \\
\bottomrule
\end{tabular}}
\end{table}

%% --- D.2 Loss Function ---
\subsection{Loss Function}

The total loss combines a reconstruction term and an isometric regularization term.
The reconstruction loss is the mean squared error between the input and the reconstructed output.
The isometric regularization, based on the Relaxed Distortion Measure, encourages the decoder mapping to locally preserve distances.
It is computed from the Gram matrix of the decoder Jacobian: the ratio of the trace of the squared Gram matrix to the square of the trace of the Gram matrix.
Latent-space augmentation is applied by interpolating (and slightly extrapolating) between pairs of encoded points:
the mixing coefficient is drawn uniformly from $[-0.2,\; 1.2)$, allowing the regularization to act beyond the convex hull of the training encodings.
The regularization weight was set to $10^{-7}$.

%% --- D.3 Training ---
\subsection{Training Hyperparameters}
\begin{table}[H]
\centering
\caption{Training hyperparameters.}
\label{tab:training}
\resizebox{0.4\linewidth}{!}{%
\begin{tabular}{l|c}
\toprule
Parameter & Value \\
\midrule
Optimizer      & Adam \\
Learning rate  & $10^{-3}$ \\
Epochs         & 1000 \\
Batch size     & 10 \\
Device         & CUDA \\
\bottomrule
\end{tabular}}
\end{table}

%% ============================================================
\vspace{1em}
\setcounter{section}{4}
\refstepcounter{section}\label{app:voo}
\noindent{\large\textbf{Appendix E. Voronoi Optimistic Optimization (VOO)}}
% \section{Voronoi Optimistic Optimization (VOO)}
% \label{app:voo}

%% --- E.1 Algorithm ---
\subsection{Algorithm}

At each iteration, VOO either performs a global uniform sample with probability $p_{\text{global}}$ or a local sample near the current best point.
For local sampling, the algorithm identifies the Voronoi radius $r$ of the best point (the squared distance to its nearest evaluated neighbor) and draws candidates within that radius.
If no valid candidate is found within $n_{\text{switch}}$ uniform trials, the sampler switches to a Gaussian centered on the best point with standard deviation $\sigma = \sigma_c \cdot \sqrt{r\,/\,d}$, where $d$ is the search-space dimensionality.
Each candidate is accepted as soon as its distance to the best point falls below the Voronoi radius, ensuring the new sample lies within the best point's Voronoi cell.

%% --- E.2 Hyperparameters ---
\subsection{Hyperparameters}
\begin{table}[H]
\centering
\caption{VOO hyperparameters used in experiments.}
\label{tab:voo-hyper}
\resizebox{0.85\linewidth}{!}{%
\begin{tabular}{l|c}
\toprule
Parameter & Value \\
\midrule
Initial samples ($n_{\text{init}}$)                  & 16  \\
Max iterations ($T$)                                  & 30  \\
Total evaluations ($n_{\text{init}} + T$)             & 46  \\
\midrule
Global sampling probability ($p_{\text{global}}$)     & 0.55 \\
Gaussian coefficient ($\sigma_c$)                     & 0.6 \\
Uniform-to-Gaussian switch ($n_{\text{switch}}$)      & 20  \\
Max inner loop                                        & 500 \\
\midrule
Search range $[x_{\min},\; x_{\max}]$                & $[-15,\; 15]$ \\
Independent runs per model                            & 10  \\
Master seed                                           & 123456789 \\
\bottomrule
\end{tabular}}
\end{table}
% Table~\ref{tab:voo-hyper} summarizes the VOO configuration shared by both settings.
% Each run uses $n_{\text{init}}=16$ uniformly drawn initial samples followed by $T=30$ VOO iterations,
% yielding $n_{\text{eval}}=46$ evaluations per run.
% The same search range $[-15,\;15]$ is used for both the latent and the raw kinematic space;
% this is appropriate because the raw features are normalized by each robot's body height (Section~\ref{app:normalization}),
% placing them on a comparable scale to the learned latent coordinates.
Each run uses $n_{\text{init}}=16$ uniformly drawn initial samples followed by $T=30$ VOO iterations,
yielding $n_{\text{eval}}=46$ evaluations per run.
The same search range $[-15,\;15]$ is used for both the latent and the raw kinematic space;
this is appropriate because the raw features are normalized by each robot's shoulder-to-base distance (Appendix~\ref{app:normalization}),
placing them on a comparable scale to the learned latent coordinates.

\end{document}